\documentclass{article}

\usepackage{PRIMEarxiv}

\usepackage[utf8]{inputenc} 
\usepackage[T1]{fontenc}    
\usepackage{hyperref}       
\usepackage{url}            
\usepackage{booktabs}       
\usepackage{amsfonts}       
\usepackage{nicefrac}       
\usepackage{microtype}      
\usepackage{lipsum}
\usepackage{fancyhdr}       
\usepackage{graphicx}       
\graphicspath{{media/}}     
\usepackage{amsmath}
\usepackage{lscape}
\usepackage{caption}
\usepackage{subcaption}
\usepackage{natbib}

\title{Towards Reliable Sea Ice Drift Estimation in the Arctic: Deep Learning Optical Flow on RADARSAT-2} 

\author{%
 Daniela Martin\\
 University of Delaware \\
 \texttt{dmartinv@udel.edu}
 \And
 Joseph Gallego \\
 Drexel University \\
 \texttt{jg3959@drexel.edu} \\
}

\begin{document}
\maketitle

\begin{abstract}
Accurate estimation of sea ice drift is critical for Arctic navigation, climate research, and operational forecasting. While \textit{optical flow}—a computer vision technique for estimating pixel-wise motion between consecutive images—has advanced rapidly in computer vision, its applicability to geophysical problems, and more specifically, to satellite SAR imagery, remains underexplored. Classical optical flow methods rely on mathematical models and strong assumptions about motion, which limit their accuracy in complex scenarios; recent deep learning-based approaches have substantially improved performance and are now the standard in computer vision, motivating their application to sea ice drift estimation. We present the first large-scale benchmark of 48 deep learning optical flow models on RADARSAT-2 ScanSAR sea ice imagery, evaluated with endpoint error (EPE) and Fl-all metrics against GNSS-tracked buoys. Several models achieve sub-kilometer accuracy (EPE 6-8 pixels, 300-400~m), a small error relative to the spatial scales of sea ice motion and typical navigation requirements in the Arctic. Our results demonstrate that the models are capable of capturing consistent regional drift patterns and that recent deep learning-based optical flow methods, which have substantially improved motion estimation accuracy compared to classical methods, can be effectively transferred to polar remote sensing. Optical flow produces spatially continuous drift fields, providing motion estimates for every image pixel rather than at sparse buoy locations, offering new opportunities for navigation and climate modeling.
\end{abstract}

\section{Introduction}
\label{sec:intro_related}
Arctic sea ice is a key component of the climate system, influencing global circulation patterns \cite{li2022impacts, willmes2023patterns}, ecosystems \cite{steiner2021climate, campbell2022monitoring}, and human activity through its role in navigation and forecasting \cite{yastrebova2021positioning, bayler2024satellite}. Accurate estimation of ice drift is therefore essential for both climate research and operational decision-making \cite{wang2022intercomparison}.  Classical approaches for sea ice motion estimation include correlation-based methods such as maximum cross-correlation and phase correlation \cite{lavergne2010sea, hollands2012motion}, as well as feature tracking strategies \cite{karvonen2012operational}. While effective in certain conditions, these methods often degrade in low-texture regions, at floe boundaries, or under strong deformation \cite{roberts2019variational, lei2021autonomous}. 

Optical flow is a computer vision technique that estimates motion at the level of individual pixels between two consecutive images. Each pixel is assigned a vector indicating the direction and magnitude of its movement from the first to the second image, providing a dense motion field across the entire scene \cite{alfarano2024estimating}. In the context of sea ice, optical flow has the potential to capture detailed drift patterns over large areas, including subtle deformations and floe interactions, which are difficult to resolve with sparse tracking methods such as drifting buoys, whose limited spatial coverage prevents the observation of fine-scale drift patterns and interactions between individual ice floes. Classical optical flow approaches rely on mathematical models, such as brightness constancy constraints combined with spatial regularization, solved through variational or local optimization methods \cite{weickert2006survey}. While effective in certain contexts, these methods are limited in handling large displacements, complex motion patterns, or noisy data. In recent years, deep learning has become the standard for computer vision tasks, including optical flow, enabling models to learn motion patterns from data and achieve state-of-the-art accuracy and efficiency \cite{shah2021traditional}.

Recent developments in optical flow estimation have focused on improving both accuracy and efficiency, extending the RAFT \cite{teed2020raft} family of methods into more scalable and generalizable architectures. WAFT \cite{wang2025waft} challenges the need for cost volumes by relying on high-resolution warping, achieving competitive accuracy with lower memory usage and faster inference. MEMFOF \cite{bargatin2025memfof} revisits RAFT-like designs with reduced correlation volumes and multi-frame estimation, striking a balance between runtime efficiency and state-of-the-art accuracy, particularly at FullHD resolution. DPFlow \cite{morimitsu2025dpflow} addresses scalability by generalizing to ultra-high-resolution inputs (up to 8K) without requiring high-resolution training, while also introducing a new benchmark to systematically evaluate methods at large scales. Complementarily, SEA-RAFT \cite{wang2024sea} improves RAFT’s convergence and generalization through a simplified architecture and novel training strategies, achieving strong cross-dataset performance with reduced computational cost. These approaches consistently achieve state-of-the-art results across standard optical flow benchmarks \cite{mehl2023spring, Butler:ECCV:2012, Geiger2012CVPR, Menze2015CVPR}. Collectively, they highlight a shift toward architectures that not only improve accuracy but also address the growing demands of resolution, efficiency, and generalization in optical flow.

Progress in this direction has been limited in part by the scarcity of large-scale labeled datasets for Arctic sea ice motion. 
The 2021 \emph{Sea Ice Dynamics Experiment (SIDEx)} campaign \cite{hutchings2023sidex,elosegui2023sidex} provides a rare opportunity by deploying a set of Global Navigation Satellite System (GNNS)  \cite{grewal2011global} tracked buoys in the Arctic pack ice, delivering high-quality ground-truth trajectories that enable rigorous benchmarking of sea ice motion estimation at scale.  

In this work, we present the first large-scale benchmark of 48 pretrained deep learning optical flow models, covering convolutional, recurrent, and transformer-based architectures, applied to RADARSAT-2 ScanSAR imagery \cite{scheuchl2004potential} over the Beaufort Sea in the Alaskan Arctic. Models performance is evaluated against the buoy positions using widely adopted metrics in the computer vision community for optical flow assessment, specifically endpoint error (EPE) and Fl-all metrics. Several achieve sub-kilometer accuracy, highlighting the promise of pretrained vision models for polar geophysical applications. Beyond quantitative scores, we report qualitative patterns and discuss implications for Arctic navigation and climate science.

\section{Experimental Setup}
\label{sec:experiment}
We evaluate the capability of deep learning optical flow methods to estimate sea ice drift from SAR imagery. 
Our experiments combine a dedicated RADARSAT-2 dataset with GNSS buoy observations, testing a wide range of pretrained models under consistent preprocessing and standardized evaluation metrics. 
This section outlines the dataset preparation, model selection, evaluation criteria, and quantitative and qualitative results.

\subsection{Dataset Preprocessing}
We use a dataset of 54 RADARSAT-2 ScanSAR images collected during the 2021 SIDEx campaign in the Beaufort Sea (March--May). The data provide wide coverage of Arctic sea ice at 20--50 m spatial resolution, encompassing diverse conditions such as compact ice, leads, and fractures. The temporal resolution of the dataset is $\approx$1.3 days.

For validation, we compare the estimated motion fields against ground-truth displacements from 43 buoys \cite{hutchings2023sidex}, including 12 SATICE buoys equipped with geodetic-quality GNSS tracking \cite{elosegui2023sidex}. The SATICE buoys recorded positions every 5 minutes, while the other buoys provided updates every 10 minutes, enabling high-temporal-resolution tracking. To align with the SAR acquisitions, we linearly interpolated buoy positions, yielding 1880 displacement observations for evaluation. To ensure consistent model input, all images were resampled to a uniform 50~m spatial resolution. Each image was then cropped to a $40~\text{km}^2$ region centered on the position of the GNSS buoy \textit{OSU-IT-23} \cite{hutchings2023sidex}. This Lagrangian frame of reference \cite{flotho2023lagrangian} allows the image patches to move with the buoy, retaining fine-scale deformation patterns.

\subsection{Optical Flow Models}
We benchmarked 48 pretrained optical flow models, spanning convolutional, recurrent, and transformer-based architectures. The selection was guided by both code availability and reproducibility of state-of-the-art optical flow methods reported in standard benchmarks \cite{mehl2023spring, Butler:ECCV:2012, Geiger2012CVPR, Menze2015CVPR}. While some top-performing models have been reported in the literature, not all had accessible or verifiable implementations. To enable consistent evaluation with our sea ice data, we included only those models for which a reliable implementation could be obtained, using pretrained weights and interfaces provided by the \textit{ptlflow} library \citep{morimitsu2021ptlflow}.

Models were applied without fine-tuning to assess their transferability to SAR imagery. While our GNSS-tracked buoy observations provide valuable ground-truth for validation, they are too sparse-i.e., they provide measurements only at a limited number of discrete locations rather than at every pixel-to reliably train or fine-tune deep learning optical flow models. Most state-of-the-art optical flow architectures are trained on dense synthetic ground-truth displacement fields, which provide per-pixel motion vectors covering nearly the entire image \cite{alfarano2024estimating}, whereas our sea ice dataset contains only a handful of buoy-based measurements. Attempting to fine-tune on this sparse data would risk overfitting to the few available observations and would also reduce the amount of data available for an independent evaluation, since a subset would have to be withheld for validation \cite{petrini2022learning}. By applying pretrained models directly, we can rigorously evaluate their out-of-the-box performance in a realistic Arctic scenario while preserving the integrity of the test set and avoiding spurious conclusions due to extremely limited training data.

Performance was quantified using two standard optical flow metrics: \textbf{Endpoint Error (EPE)}, the Euclidean distance between predicted and reference displacements (reported in pixels and converted to meters), and \textbf{Flow outlier rate (Fl-all)}, the proportion of pixels with error $>3$ pixels and $>5\%$ relative displacement. These metrics capture both absolute error and robustness to large displacements, providing complementary perspectives on model performance.

\section{Results and Discussions}
\textbf{Quantitative Analysis.} 75\% of the tested models achieved EPE values within 6--8 pixels, indicating consistent performance across the majority of models. Table~\ref{tab:results_top5} presents the top five models, ranked by Fl-all to emphasize robustness against gross displacement errors in operationally relevant regions. Full quantitative results for all 48 models are provided in the supplemental material.

\begin{table}[htbp]
\centering
\caption{Top-5 optical flow models evaluated on RADARSAT-2 ScanSAR imagery. Models are ranked by Fl-all, which quantifies the fraction of pixels with gross displacement errors, prioritizing reliability in operationally sensitive regions. EPE is reported in pixels (50 m/pixel). Lower values indicate better performance in both average error and robustness against gross errors. Best value for each metric is highlighted in bold.}
\label{tab:results_top5}
\begin{tabular}{lcc}
\toprule
Model & EPE [px] & Fl-all [\%] \\
\midrule
DIP (Sintel) \cite{zheng2022dip}& 6.29 & \textbf{42.52} \\
RPKNet (Sintel) \cite{morimitsu2024recurrent}& 7.51 & 42.89 \\
SEA-RAFT (M) (Spring) \cite{wang2024sea}& 7.56 & 43.07 \\
DPFlow (Chairs) \cite{morimitsu2025dpflow}& \textbf{6.27} & 43.08 \\
SEA-RAFT (S) (Spring) \cite{wang2024sea}& 7.75 & 43.12 \\
\bottomrule
\end{tabular}
\end{table}

\begin{figure}[htbp]
    \centering
    \includegraphics[width=0.7\textwidth]{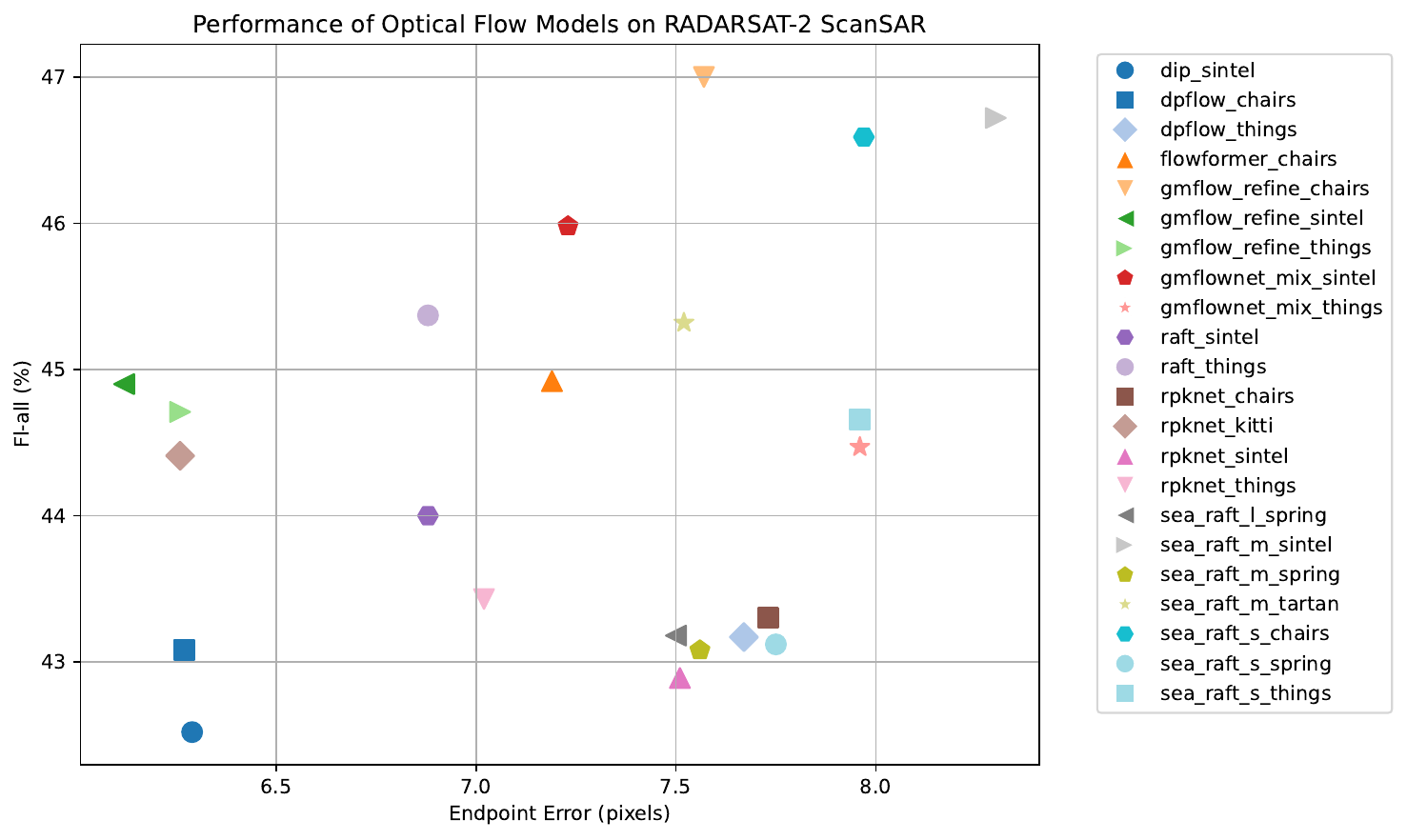}
    \caption{Scatter plot of optical flow models evaluated on RADARSAT-2 ScanSAR imagery, comparing end-point error (EPE) and flow outlier rate (Fl-all). For clarity, only models with EPE < 8.5 and Fl-all < 47 are shown (22 models in total). A total of 48 models were evaluated (full list in Table~\ref{tab:all_results}), but the full set is omitted here to avoid overcrowding. For comparison, the benchmark metrics officially reported in the PTLFlow library for standard computer vision datasets are summarized in Table~\ref{tab:summarized_metrics_benchmarks}.}
    \label{fig:epe_flall_scatter}
\end{figure}

Figure~\ref{fig:epe_flall_scatter} presents a scatterplot of the best performing 22 optical flow models, showing EPE versus Fl-all. To improve readability, we only visualize models within a restricted range of errors. Specifically, we plot models with end-point error (EPE) below 8.5 and flow outlier rate (Fl-all) below 47. This filtering allows us to highlight the most competitive methods, since displaying all 48 evaluated models would overcrowd the figure and obscure relevant patterns.

\begin{figure*}[t]
    \centering
    \begin{subfigure}{0.23\textwidth}
        \centering
        \includegraphics[width=\linewidth]{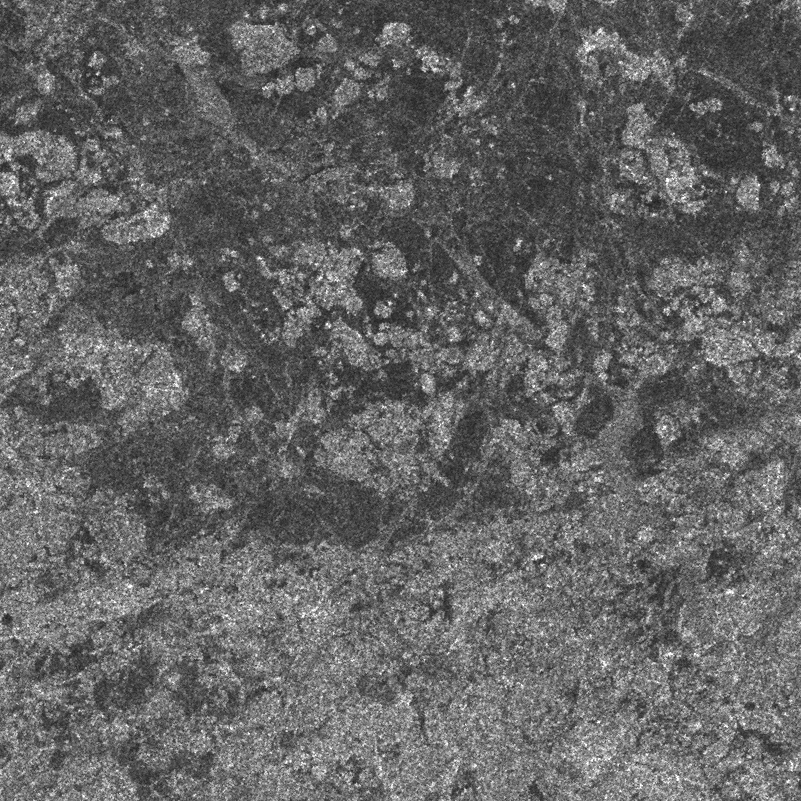}
        \subcaption{\vspace{3.36mm}}
    \end{subfigure}
    \hfill
    \begin{subfigure}{0.23\textwidth}
        \centering
        \includegraphics[width=\linewidth]{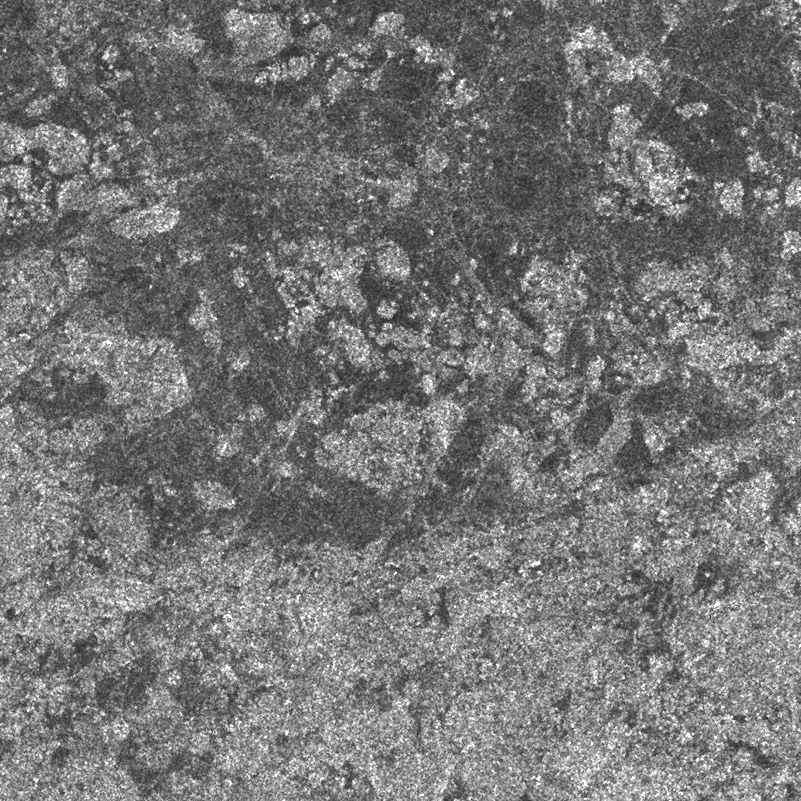}
        \subcaption{\vspace{3.36mm}}
    \end{subfigure}
    \hfill
    \begin{subfigure}{0.23\textwidth}
        \centering
        \includegraphics[width=\linewidth]{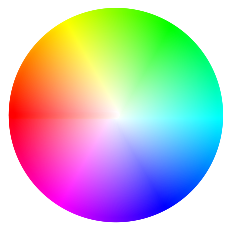}
        \subcaption{\vspace{3.36mm}}
    \end{subfigure}
    \vfill
    \begin{subfigure}{0.23\textwidth}
        \centering
        \includegraphics[width=\linewidth]{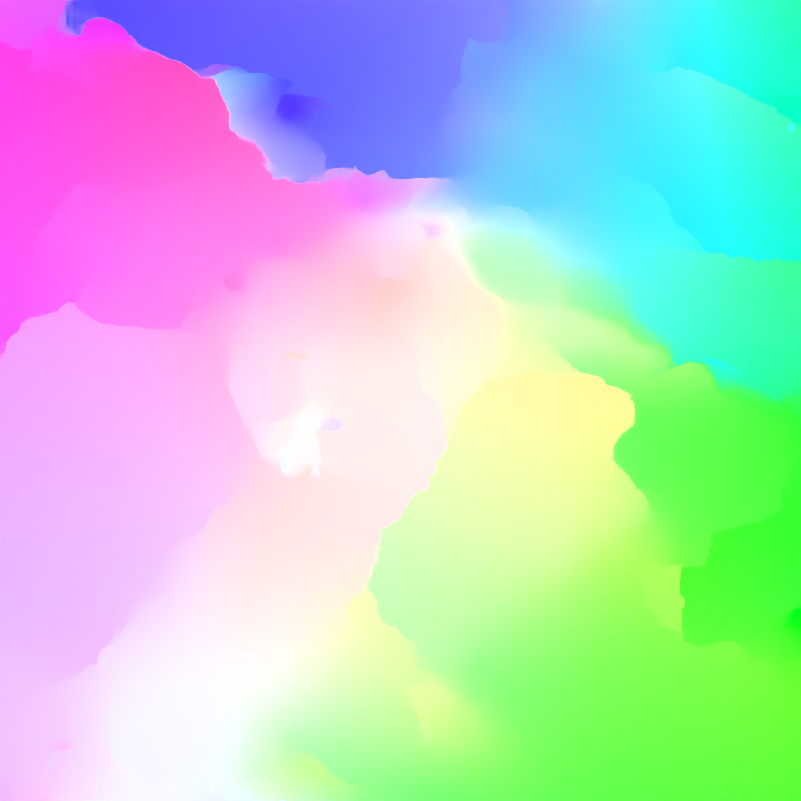}
        \subcaption{\vspace{3.36mm}}
    \end{subfigure}
    \hfill
    \begin{subfigure}{0.23\textwidth}
        \centering
        \includegraphics[width=\linewidth]{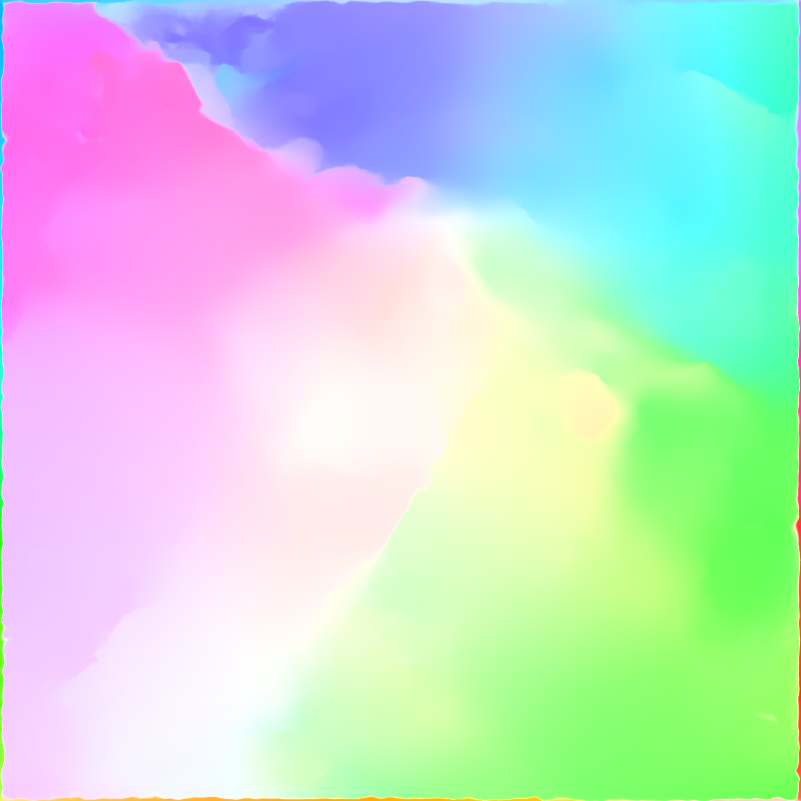}
        \subcaption{\vspace{3.36mm}}
    \end{subfigure}
    \hfill
    \begin{subfigure}{0.23\textwidth}
        \centering
        \includegraphics[width=\linewidth]{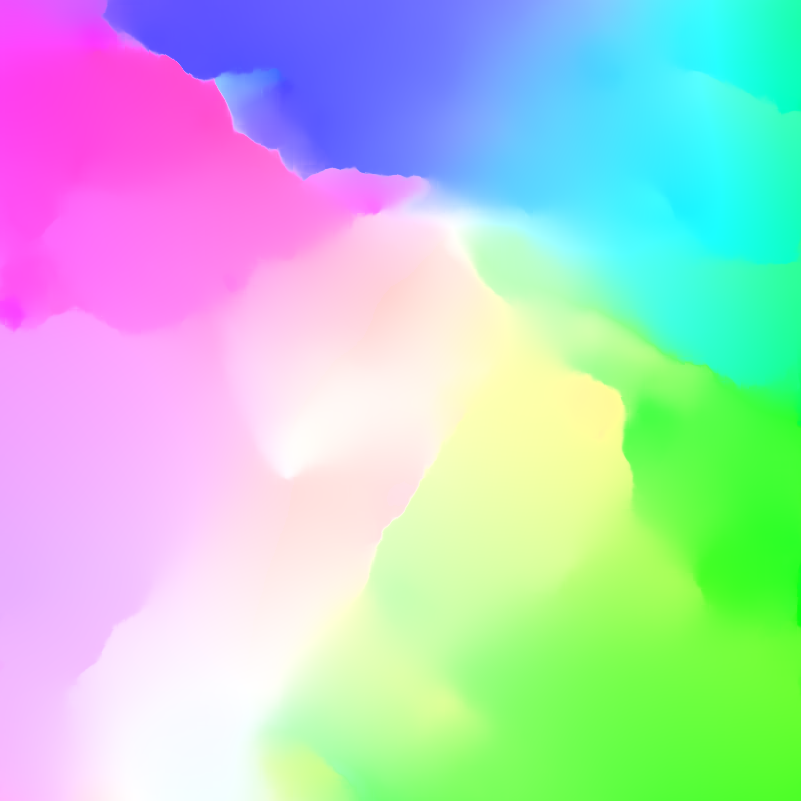}
        \subcaption{\vspace{3.36mm}}
    \end{subfigure}
    \caption{Qualitative comparison of the top three models on a selected RADARSAT-2 ScanSAR image pair. (a) First image. (b) Second image. (c) Colorwheel for interpretation: hue indicates direction, saturation indicates magnitude. Magnitude increases radially from the center (0) to the maximum displacement in the dataset ($\approx$100 pixels for this case) (d) DIP (Sintel) prediction \cite{zheng2022dip}. (e) RPKNet (Sintel) prediction \cite{morimitsu2024recurrent}. (f) SEA-RAFT (M) (Spring) prediction \cite{wang2024sea}.}
    \label{fig:qualitative}
\end{figure*}

\textbf{Qualitative Analysis.} To further assess model performance, we conducted a qualitative analysis on a representative RADARSAT-2 ScanSAR image pair. Figure~\ref{fig:qualitative} illustrates the predictions from the three best-performing models. It can be interpreted by referring to the standard optical flow colorwheel (subfigure c), where hue indicates direction and saturation indicates magnitude.

Across all models (d–f), several distinct drift blocks can be identified, each characterized by a dominant color representing a coherent motion direction. The magnitude of motion increases gradually toward the corners of the scene and is lowest near the image center, an expected pattern given that the analysis is performed in a Lagrangian reference frame. All three models show a strong agreement in motion direction, capturing consistent regional drift pathways; however, differences emerge in the estimated magnitude.

The RPKNet model (e) systematically underestimates the motion magnitude compared to DIP (d) and SEA-RAFT (f), particularly in areas where multiple motion blocks converge. In addition, RPKNet produces spurious high-gradient values along the image borders that do not correspond to physically feasible ice motion. For practical use, we recommend cropping 5--10 pixels along the borders when applying this model to operational scenarios. In contrast, DIP and SEA-RAFT yield more stable and spatially coherent results, better preserving the large-scale drift structure.

For Arctic applications such as navigation support and early-warning systems, models that produce less nuanced and more categorical motion estimates---favoring stable regional drift patterns over subtle local variations---may be more reliable and interpretable. Overall, the qualitative results confirm that deep learning optical flow models can capture realistic sea ice motion fields, while highlighting the need for model-specific considerations depending on the intended operational context.

\textbf{Discussion.} Our large-scale benchmark of 48 deep learning optical flow models on RADARSAT-2 ScanSAR data demonstrates that pretrained computer vision methods can be effectively transferred to polar remote sensing. Remarkably, 75\% of models achieved endpoint errors (EPE) within 6-8 pixels (approximately 300-400 m), with several methods surpassing this level of accuracy. Notably, Fl-all scores are relatively high because this metric is extremely strict, designed for contexts requiring subpixel precision; nevertheless, we focused on selecting the best values to minimize large errors as much as possible. While such errors may seem substantial in a general computer vision context, at Arctic scales they are negligible, since sea ice drift patterns evolve over tens of kilometers, making pixel-level deviations minor in the context of regional dynamics. The consistency across diverse architectures suggests that adaptations designed for complex textures and varying illumination in natural images also translate well to SAR imagery. 

When compared to standard computer vision benchmarks (Table~\ref{tab:summarized_metrics_benchmarks}), the EPE values obtained for RADARSAT-2 imagery are higher in absolute terms. However, this discrepancy is expected given the fundamental differences between the domains: SAR imagery exhibits distinct speckle noise patterns and limited texture gradients, ground-truth motion fields are sparse and less precise, and the physical scales of motion are orders of magnitude larger. Therefore, while the reported errors may seem high relative to benchmarks such as Sintel or KITTI, they are operationally negligible in Arctic applications, where ice drift occurs over tens of kilometers.

Qualitative analysis further showed that all models largely agree on motion patterns, although they may differ in edge definition and slightly in motion magnitude, while the motion direction is highly consistent. These qualitative observations further highlight subtle model-dependent trade-offs between smoothness, sharpness, and magnitude estimation that are consistent with the quantitative metrics reported above.

\textbf{Limitations.} A key limitation remains the lack of large, labeled SAR datasets for supervised training and fine-tuning. Manually generating ground truth fields at pixel level is nearly infeasible, making pretrained models a key ally in the near term.

\section{Conclusion}
Our benchmark provides the first comprehensive evidence that pretrained deep learning optical flow models can deliver operationally meaningful accuracies for sea ice drift estimation from SAR. These findings demonstrate the promise of transferring vision-based methods to geophysical applications and chart a path forward for developing domain-specific models tailored to the unique challenges of polar remote sensing.

Minor qualitative differences between architectures suggest that targeted adaptations to SAR-specific characteristics could further enhance accuracy and stability. Future progress will likely require domain-adapted approaches that combine machine learning with physical constraints, exploit multi-sensor fusion, and extend to long temporal sequences for seasonal-scale monitoring. Moreover, future work could explore the use of alternative evaluation metrics better suited to geophysical problems that capture spatial and temporal dependencies more effectively.


\section*{Acknowledgements}
The authors acknowledge the U.S. National Ice Center for providing RADARSat-2 imagery used in this study. Ground truth data were provided by multiple collaborating institutions, as cited in the text. Due to data usage restrictions, the satellite imagery cannot be publicly shared.

\bibliographystyle{unsrt}  
\bibliography{references}  

\appendix

\section{Appendix / supplemental material}

\subsection{Extended Background} \label{sec:supp_background}
\subsubsection{Classical Methods for Sea Ice Motion Estimation}
Traditional sea ice motion estimation techniques rely on block matching, cross-correlation, and feature tracking applied to satellite images \cite{hollands2012motion}. Maximum cross-correlation (MCC) \cite{lavergne2010sea} and phase correlation \cite{xian2017super} have been widely adopted due to their efficiency. Extensions include Laplacian pyramids \cite{haarpaintner2004automatic} and wavelet-based methods \cite{zhao2001principal}, aimed at improving robustness. However, all remain limited in regions of thin ice, low texture, or under rapid deformation \cite{meng2024synthetic}. Sparse methods based on tie points \cite{karvonen2012operational} offer reliability in structured regions but fail to capture full ice dynamics, especially under divergence or ridging \cite{roberts2019variational}. Dense optical flow offers a richer representation \cite{vogel2012optical} but must overcome SAR-specific noise.

\subsubsection{Deep Learning for Optical Flow}
Deep learning introduced end-to-end frameworks for motion estimation, beginning with FlowNet \cite{dosovitskiy2015flownet} and FlowNet2 \cite{ilg2017flownet}. PWC-Net \cite{sun2018pwc} introduced feature pyramids and cost volumes, while RAFT \cite{teed2020raft} advanced accuracy through recurrent refinement. More recently, transformer-based models such as Unimatch \cite{xu2023unifying} and others exploit global feature matching and self-attention, capturing long-range dependencies in motion. These advances have shown strong performance in RGB benchmarks, but their transfer to SAR remains largely untested \cite{meng2024synthetic}.  

\subsubsection{Applications in Climate Science and Navigation}
Satellite-derived ice motion fields inform large-scale climate models of Arctic dynamics \cite{sandven2023sea}, while operational services (e.g., NIC, Copernicus Marine Service) use them to improve forecasts and navigation safety. Enhanced methods also support search and rescue, oil spill tracking, and polar field campaigns \cite{malik2024climate}. Improving motion estimation with deep learning could therefore benefit both scientific and operational communities.

\subsection{Study area}
\begin{figure}[htbp]
    \centering
    \includegraphics[width=0.8\linewidth]{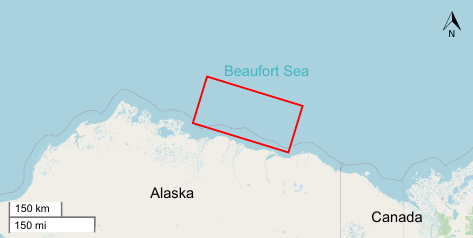}
    \caption{Study area in the Beaufort Sea, Alaska, showing the region of interest used for benchmarking optical flow models.}
    \label{fig:study_area}
\end{figure}

Fig. \ref{fig:study_area} shows the study area in the Beaufort Sea, Alaska.

\subsection{Dataset Details}
The RADARSAT-2 ScanSAR dataset spans March 4--May 16, 2021, comprising 54 images at 20, 40, and 50 m spatial resolution. The dataset captures diverse Arctic sea ice states, including compact floes, divergence zones, and lead formation.

\subsection{Preprocessing Pipeline}
Images were orthorectified, co-registered using satellite ephemeris data, and normalized to mitigate SAR-specific intensity variations.

\subsection*{Results Across All Models}
While the main paper highlights the top five models, Table~\ref{tab:all_results} reports full results for all 48 evaluated architectures. This provides a comprehensive reference for future studies comparing deep optical flow methods in geophysical contexts.

\begin{table}[htbp]
\centering

\caption{Endpoint Error (EPE) mean and standard deviation, and Fl-all mean and standard deviation for all 48 evaluated models.}
\begin{tabular}{lcccc}
\toprule
Model Name & EPE Mean [px] & EPE Std [px] & Fl-all Mean [\%] & Fl-all Std [\%] \\
\midrule
dip\_kitti \cite{zheng2022dip} & 8.04 & 8.98 & 51.3 & 33.67 \\
dip\_sintel \cite{zheng2022dip} & 6.29 & 5.09 & 42.52 & 30.06 \\
dip\_things \cite{zheng2022dip} & 60.15 & 127.68 & 49.24 & 34.36 \\
dpflow\_chairs \cite{morimitsu2025dpflow} & 6.27 & 5.3 & 43.08 & 29.83 \\
dpflow\_kitti \cite{morimitsu2025dpflow} & 8.55 & 9.41 & 63.13 & 34.72 \\
dpflow\_sintel \cite{morimitsu2025dpflow} & 7.76 & 6.88 & 53.07 & 37.25 \\
dpflow\_things \cite{morimitsu2025dpflow} & 7.67 & 11.55 & 43.17 & 30.75 \\
flowformer\_chairs \cite{huang2022flowformer} & 7.19 & 7.98 & 44.92 & 31.13 \\
gmflow\_chairs \cite{xu2022gmflow} & 8.35 & 11.63 & 55.88 & 36.24 \\
gmflow\_kitti \cite{xu2022gmflow} & 7.21 & 6.48 & 55.74 & 31.57 \\
gmflow\_refine\_chairs \cite{xu2022gmflow} & 7.57 & 10.66 & 47 & 31.92 \\
gmflow\_refine\_kitti \cite{xu2022gmflow} & 6.85 & 6.09 & 54.38 & 30.33 \\
gmflow\_refine\_sintel \cite{xu2022gmflow} & 6.12 & 5.36 & 44.9 & 30.53 \\
gmflow\_refine\_things \cite{xu2022gmflow} & 6.26 & 5.27 & 44.71 & 29.4 \\
gmflow\_sintel \cite{xu2022gmflow} & 7.62 & 9.57 & 54.05 & 33.84 \\
gmflow\_things \cite{xu2022gmflow} & 9.02 & 18.45 & 52.38 & 33.89 \\
gmflownet\_kitti \cite{zhao2022global} & 8.04 & 7.15 & 53.43 & 35.19 \\
gmflownet\_mix\_sintel \cite{zhao2022global} & 7.23 & 7.53 & 45.98 & 32.27 \\
gmflownet\_mix\_things \cite{zhao2022global} & 7.96 & 13.29 & 44.47 & 31.09 \\
neuflow2\_mixed \cite{zhang2024neuflow} & 7.43 & 8.6 & 47.41 & 32.5 \\
pwcnet\_things \cite{sun2018pwc} & 8.54 & 16.5 & 44.76 & 28.94 \\
raft\_kitti \cite{teed2020raft} & 9.25 & 11.37 & 55.86 & 34.28 \\
raft\_sintel \cite{teed2020raft}  & 6.88 & 7.04 & 44 & 31.15 \\
raft\_things \cite{teed2020raft}  & 6.88 & 6.25 & 45.37 & 31.71 \\
rpknet\_chairs \cite{wang2024sea} & 7.73 & 12.01 & 43.3 & 30.56 \\
rpknet\_kitti \cite{wang2024sea} & 6.26 & 5.32 & 44.41 & 30.79 \\
rpknet\_sintel \cite{wang2024sea} & 7.51 & 11.33 & 42.89 & 30.34 \\
rpknet\_things \cite{wang2024sea} & 7.02 & 7.96 & 43.43 & 30.82 \\
sea\_raft\_l\_chairs \cite{wang2024sea} & 17.15 & 19.99 & 47.55 & 30.57 \\
sea\_raft\_l\_kitti \cite{wang2024sea} & 17.81 & 13.6 & 86.19 & 24.53 \\
sea\_raft\_l\_sintel \cite{wang2024sea} & 8.51 & 12.75 & 45.89 & 33.02 \\
sea\_raft\_l\_spring \cite{wang2024sea} & 7.5 & 10.66 & 43.18 & 30.39 \\
sea\_raft\_l\_tartan \cite{wang2024sea} & 7.84 & 11.54 & 47.37 & 32.93 \\
sea\_raft\_l\_things \cite{wang2024sea} & 126.52 & 169.42 & 61.53 & 39.04 \\
sea\_raft\_m\_chairs \cite{wang2024sea} & 8.85  & 10.16 & 45.33 & 30.92 \\
sea\_raft\_m\_kitti \cite{wang2024sea}  & 12.2  & 12.33 & 82.35 & 27.84 \\
sea\_raft\_m\_sintel \cite{wang2024sea} & 8.3   & 12.26 & 46.72 & 32.89 \\
sea\_raft\_m\_spring \cite{wang2024sea} & 7.56  & 11.09 & 43.08 & 30.34 \\
sea\_raft\_m\_tartan \cite{wang2024sea} & 7.52  & 10.91 & 45.32 & 31.01 \\
sea\_raft\_m\_things \cite{wang2024sea} & 18.11 & 36.08 & 47.89 & 34.11 \\
sea\_raft\_s\_chairs \cite{wang2024sea} & 7.97  & 11.16 & 46.59 & 32.31 \\
sea\_raft\_s\_kitti \cite{wang2024sea}  & 12.4  & 13.71 & 63.83 & 34.47 \\
sea\_raft\_s\_sintel \cite{wang2024sea} & 9.36  & 10.87 & 68.44 & 28.93 \\
sea\_raft\_s\_spring \cite{wang2024sea} & 7.75  & 12.26 & 43.12 & 30.34 \\
sea\_raft\_s\_things \cite{wang2024sea} & 7.96  & 11.29 & 44.66 & 32.03 \\
unimatch\_things \cite{xu2023unifying}     & 9.02  & 18.45 & 52.38 & 33.89 \\
waft\_chairs \cite{wang2025waft}         & 8.87  & 10.75 & 52.92 & 35.54 \\
waft\_things \cite{wang2025waft}         & 10.03 & 25.67 & 49.59 & 34.45
\end{tabular}%
\label{tab:all_results}
\end{table}

\subsection*{Reference Metrics from the PTLFlow Library}
Table~\ref{tab:summarized_metrics_benchmarks} compiles the publicly available evaluation metrics (EPE and Fl-all) from the PTLFlow library \cite{morimitsu2021ptlflow}, covering a wide range of optical flow architectures. 
These values correspond to results on standard computer vision benchmarks and are included here to provide a reference context for the models evaluated in this study.
The metrics in this table are not directly comparable to those reported for the RADARSAT-2 benchmark in this study, due to differences in data modality (RGB vs. SAR), ground-truth density, and spatial scale of motion.

\begin{landscape}
\begin{table}[htbp]
\centering
\caption{Summary of end-point error (EPE) and flow outlier rate (Fl-all) metrics available in the PTLFlow library \cite{morimitsu2021ptlflow}. Only metrics officially reported by the library are included. These results provide a reference for the relative performance of standard optical flow models under common computer vision benchmarks.}
\scalebox{0.9}{%
\begin{tabular}{llcccccccc}
\toprule
model              & checkpoint & sintel-clean-epe & sintel-clean-flall & sintel-final-epe & sintel-final-flall & kitti-2012-epe & kitti-2012-flall & kitti-2015-epe & kitti-2015-flall \\
\midrule
dip                & kitti      & 1.176                    & 3.164                      & 2.168                    & 6.721                      & 0.945                  & 2.861                    & 1.129                  & 3.424                    \\
dip                & sintel     & 0.504                    & 2.037                      & 1.494                    & 5.506                      & 1.619                  & 5.922                    & 3.977                  & 11.139                   \\
dip                & things     & 1.332                    & 3.414                      & 3.187                    & 7.939                      & 1.74                   & 6.918                    & 4.426                  & 13.349                   \\
flowformer         & chairs     & 1.952                    & 6.208                      & 4.156                    & 10.687                     & 4.217                  & 23.041                   & 9.182                  & 33.215                   \\
gmflow             & chairs     & 3.225                    & 15.873                     & 4.432                    & 18.904                     & 8.623                  & 47.849                   & 16.542                 & 54.489                   \\
gmflow             & kitti      & 2.858                    & 11.174                     & 3.906                    & 14.48                      & 2.077                  & 9.265                    & 2.104                  & 7.889                    \\
gmflow\_refine  & chairs     & 2.305                    & 7.214                      & 3.637                    & 10.668                     & 5.636                  & 23.869                   & 13.161                 & 37.014                   \\
gmflow\_refine  & sintel     & 0.762                    & 2.414                      & 1.11                     & 4.279                      & 1.638                  & 6.557                    & 1.975                  & 7.728                    \\
gmflow\_refine  & things     & 1.084                    & 3.652                      & 2.474                    & 7.331                      & 3.132                  & 13.533                   & 6.747                  & 21.673                   \\
gmflow\_refine  & kitti      & 2.443                    & 8.395                      & 3.645                    & 12.582                     & 1.59                   & 6.479                    & 1.363                  & 5.119                    \\
gmflownet          & things     & 1.309                    & 3.908                      & 2.727                    & 7.827                      & 1.963                  & 8.418                    & 4.658                  & 15.837                   \\
gmflownet          & kitti      & 4.369                    & 10.123                     & 6.239                    & 14.357                     & 1.207                  & 4.001                    & 0.621                  & 1.502                    \\
neuflow2           & mixed      & 0.999                    & 3.733                      & 1.661                    & 6.513                      & 1.293                  & 4.691                    & 1.666                  & 6.106                    \\
pwcnet             & things     & 2.643                    & 8.811                      & 4.059                    & 12.592                     & 4.042                  & 20.243                   & 10.288                 & 33.014                   \\
raft               & things     & 1.458                    & 4.173                      & 2.684                    & 7.721                      & 2.135                  & 9.257                    & 5.018                  & 17.046                   \\
raft               & sintel     & 0.763                    & 2.855                      & 1.218                    & 4.936                      & 1.338                  & 4.587                    & 1.555                  & 5.573                    \\
raft               & kitti      & 4.544                    & 10.511                     & 6.157                    & 14.794                     & 1.283                  & 4.272                    & 0.643                  & 1.58                     \\
rpknet             & chairs     & 2.406                    & 7.341                      & 3.588                    & 10.697                     & 3.965                  & 22.812                   & 10.878                 & 33.835                   \\
rpknet             & kitti      & 1.478                    & 4.605                      & 2.37                     & 7.86                       & 1.171                  & 3.817                    & 0.946                  & 2.904                    \\
rpknet             & sintel     & 0.552                    & 2.19                       & 0.813                    & 3.659                      & 1.192                  & 4.112                    & 1.257                  & 4.273                    \\
rpknet             & things     & 1.12                     & 3.516                      & 2.449                    & 7.08                       & 1.674                  & 6.817                    & 3.783                  & 12.998                   \\
sea\_raft\_l & spring     & 1.197                    & 3.074                      & 2.207                    & 6.131                      & 2.342                  & 7.856                    & 4.371                  & 12.424                   \\
sea\_raft\_l & kitti      & 0.617                    & 2.025                      & 0.931                    & 3.324                      & 1.049                  & 3.219                    & 0.817                  & 2.267                    \\
sea\_raft\_l & things     & 1.226                    & 3.116                      & 3.444                    & 7.053                      & 2.061                  & 7.898                    & 3.723                  & 12.377                   \\
sea\_raft\_l & chairs     & 2.441                    & 5.422                      & 9.272                    & 11.951                     & 5.292                  & 31.121                   & 10.5                   & 37.727                   \\
sea\_raft\_l & tartan     & 2.316                    & 7.701                      & 5.549                    & 15.499                     & 1.276                  & 4.205                    & 3.306                  & 9.977                    \\
sea\_raft\_l & sintel     & 0.444                    & 1.597                      & 0.613                    & 2.292                      & 1.086                  & 3.391                    & 0.892                  & 2.571                    \\
sea\_raft\_m & tartan     & 2.366                    & 6.752                      & 5.203                    & 14.289                     & 1.385                  & 4.467                    & 3.568                  & 9.887                    \\
sea\_raft\_m & chairs     & 2.27                     & 5.382                      & 5.744                    & 11.259                     & 4.359                  & 25.483                   & 9.159                  & 33.925                   \\
sea\_raft\_m & things     & 1.272                    & 3.235                      & 3.853                    & 7.343                      & 1.971                  & 7.406                    & 4.293                  & 13.935                   \\
sea\_raft\_m & sintel     & 0.434                    & 1.573                      & 0.584                    & 2.197                      & 1.125                  & 3.538                    & 0.88                   & 2.551                    \\
sea\_raft\_m & kitti      & 0.646                    & 2.136                      & 0.958                    & 3.49                       & 1.078                  & 3.373                    & 0.808                  & 2.213                    \\
sea\_raft\_m & spring     & 1.389                    & 3.436                      & 2.3                      & 6.455                      & 2.869                  & 8.532                    & 5.812                  & 14.49                    \\
sea\_raft\_s & chairs     & 2.052                    & 5.406                      & 4.711                    & 10.625                     & 4.401                  & 27.262                   & 9.579                  & 36.224                   \\
sea\_raft\_s & things     & 1.283                    & 3.34                       & 3.729                    & 7.307                      & 2.001                  & 8.172                    & 4.433                  & 14.694                   \\
sea\_raft\_s & sintel     & 0.547                    & 1.92                       & 0.783                    & 3.009                      & 1.153                  & 3.754                    & 1.054                  & 3.151                    \\
sea\_raft\_s & kitti      & 1.067                    & 3.072                      & 1.756                    & 5.553                      & 1.093                  & 3.589                    & 0.935                  & 2.7                      \\
sea\_raft\_s & spring     & 1.697                    & 4.064                      & 2.573                    & 7.095                      & 3.552                  & 10.453                   & 7.906                  & 18.163                  
\end{tabular}
}
\label{tab:summarized_metrics_benchmarks}
\end{table}
\end{landscape}

\end{document}